%% file: main.tex
\documentclass[a4, 10 pt, conference]{ieeeconf}  

\IEEEoverridecommandlockouts                              

\usepackage{graphicx} 
\usepackage{subcaption}
\usepackage{caption}
\usepackage{comment}
\usepackage{multirow}
\usepackage{url}

\title{\LARGE \bf
A Dialogue Robot System to Improve Credibility \\ in Sightseeing Spot Recommendations
}

\author{
Naoki Yoshimaru$^{1}$, Tomohiro Masuda$^{2}$, Hyejin Hong$^{2}$, Yusei Tanaka$^{2}$, Motoharu Okuma$^{2}$, \\ Nagihiro Matsumoto$^{2}$, Kazuma Kusu$^{1}$, Takamasa Iio$^{2}$, and Kenji Hatano$^{2}$ 
\thanks{$^{1}$ Graduate School of Culture and Information Science, Doshisha University, 1-3 Tatara-Miyakodani
Kyotanabe, Kyoto 610-0394, Japan}%
\thanks{$^{2}$ Faculty of Culture and Information Science,Doshisha University, 1-3 Tatara-Miyakodani
Kyotanabe, Kyoto 610-0394, Japan}%
}

\begin{document}

\maketitle
\thispagestyle{empty}
\pagestyle{empty}

\begin{abstract}
Various studies have been conducted on human-supporting robot systems. These systems have been put to practical use over the years and are now seen in our daily lives. In particular, robots communicating smoothly with people are expected to play an active role in customer service and guidance. In this case, it is essential to determine whether the customer is satisfied with the dialog robot or not. However, it is not easy to satisfy all of the customer’s requests due to the diversity of the customer’s speech. In this study, we developed a dialog mechanism that prevents dialog breakdowns and keeps the customer satisfied by providing multiple scenarios for the robot to take control of the dialog. We tested it in a travel destination recommendation task at a travel agency.
\end{abstract}

\section{INTRODUCTION}
\label{s:intro}
\input{tex/010-introduction}

\section{CISRobot Architecture} \label{s:archi}
\input{tex/020-architecture}

\subsection{Conversation Flow} \label{ss:convflow}
\input{tex/021-conversation-flow}


\section{CISRobot's Behavior Definition} \label{s:definition}
\input{tex/030-cisrobot-definition}

\section{Field Trial} \label{s:exper}

\input{tex/040-experiment}

\subsection{Competition} \label{ss:results}
\input{tex/041-competition-result}

\subsection{Result} \label{ss:analysis}
\input{tex/042-analysis}

\subsection{Discussion} \label{ss:discussion}
\input{tex/043-discussion}

\section{Conclusion} \label{s:concl}

\input{tex/050-conclusion}

\section*{ACKNOWLEDGMENT}
This research was funded by JSPS KAKENHI Grant Number JP19H05691 and JP19H01138.


\bibliographystyle{IEEEtran}


\end{document}

%% file: tex/010-introduction.tex
A dialogue robot competition is being held to test multimodal dialogue techniques using humanoid robots\cite{DRC2021}. 
In that dialog robot competition, a humanoid robot acts as a counter-sales person for a travel agency and recommends travel destinations to users. 
Participants compete in terms of user satisfaction with the robot's recommendations and the naturalness of the dialogue. 
In order to improve user satisfaction, it is necessary to not only present information on travel destinations but also to listen to user requests and recommend destinations that meet those requests.

It is not easy to listen to user requests through dialogue, respond to those requests, and reflect those requests in the recommendations. 
In particular, when users are allowed to speak freely, classical rule-based dialogue systems often break down. This is because the content of user utterances is diverse, and it is impossible to prepare rules for all utterances. The breakdown of dialogue can be avoided by processing the occurrence of utterances that do not exist in the rules(exception handling). However, frequent occurrences of exceptions may reduce user satisfaction. These problems also arise in example-based dialogue systems. Another approach is to use a generation-based dialog system. Generative dialogue systems can generate natural responses using large-scale language models and prompts. However, for the specific task of recommending destinations, they may generate information that is not true (fake information). For example, the system may output information about exhibits or products that do not exist in the destination description. Since such fake information may cause ethical and legal problems, adopting this method with current technology is not easy. Therefore, this study adopts a rule-based dialogue system where the designer has complete control over the utterances, although it is not as versatile as a generation-based dialogue system.

In a rule-based dialog system, the robot must take the initiative in the dialog to reflect the user's requests in the recommendations. Specifically, the robot takes the initiative in the dialog robot by repeatedly asking the user questions. Here, the robot's questions should be questions that can be answered with Yes/No or choices (choice-type questions). For example, "Do you travel alone? Or do you travel with your family or friends?." These questions should be designed so that the user can answer them. Such questions implicitly limit how users respond, making it easier to predict their answers. As a result, it can reduce the possibility of dialogue breakdown while still allowing the user to speak regarding the destination recommendation.

However, in recommending travel destinations, the user is not likely to be fully satisfied if the robot takes the initiative in the dialog and asks many questions. 
Methods in which the robot takes the initiative in dialog have been applied to daily conversations with the elderly \cite{iio2020twin}, and dialog with visitors at events \cite{iio2017retaining}. 
According to those studies, the method can elicit some degree of user satisfaction. However, the dialogue in those tasks is closer to chit-chat and different from the dialogue in recommending travel destinations. Being bombarded with questions by the robot in a chat is, in a sense, easier for the user, who has no purpose in interacting with the robot. This is because the user does not have to think about what to say to the robot. On the other hand, in recommending a travel destination, the user has the explicit goal of selecting a destination. Therefore, it is important that the robot's questions are meant for destination selection and that the user's answers are reflected in the final destination recommendation.

In this study, therefore, a dialog system was developed in which the robot takes the initiative in the dialog and, in addition to the method of specifying the customer's requests with choice-type questions, a mechanism was introduced to memorize the user's speech and reflect the content of the user's speech in the recommendations. The objective is to show through a dialog robot competition 2022 \cite{DRC2022} (DRC2022) what extent customers are satisfied with the system and to what extent they are receptive to the destinations recommended by the robot.

%% file: tex/020-architecture.tex
DRC2022 imposes a recommendation task on participants of the competition, that is to recommend the one selected by the operator in two sightseeing spots chosen by a participant.
Moreover, the limitation of time for the recommendation task is for five minutes.
Hence, a robot for DRC2022 has to collect information about a participant and trip purposes in a limited time.
Therefore, we develop a framework for managing a flow of dialogues and equip into our robot named CISRobot.

This section describes an architecture of our framework named ``Conversation Flow,'' and does generalized functions in our framework for enabling us to realize every dialogue combined with the robot's behaviors.
Moreover, we also describe common behaviors of CISRobot implemented to increase user's impression of our robot during conversations.

%% file: tex/021-conversation-flow.tex
We aim to design CISRobot enabling us to certainly perform the recommendation task.
Hence, to complete the task, a robot needs to collect information about a dialogue partner and some purposes of their trip in a limited time. 
Our framework ``Conversation Flow (CF)'' can manage a flow of conversations constructed by monologues and dialogues. 
A robot speaks according to the contents of predefined monologues and dialogues so that the robot certainly listens to information about a participant and the purposes of their trip from users. 

\begin{figure}[t]
    \centering
    \includegraphics[width=0.8\hsize]{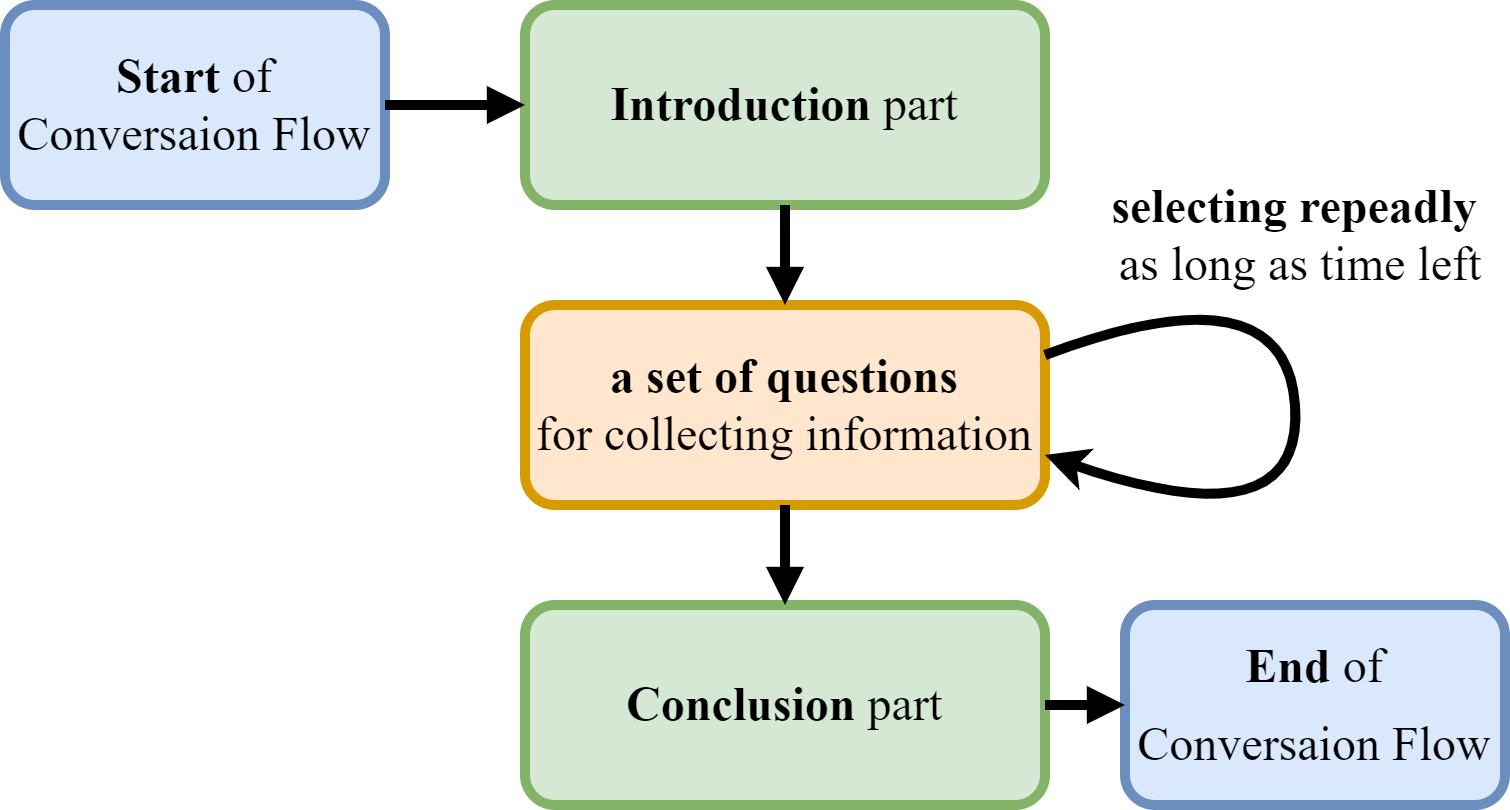}
    \caption{{\bfseries A framework of our conversation flow: }
        We design a flow of conversations in this figure.
        Part {\bfseries 2.} for collecting participant's information is a set of questions and monologues defined by a robot administrator in advance.
    }
    \label{fig:cf-flow}
\end{figure}

A CF is a series of three parts as follows:
\begin{description}
    \item [\bfseries 1. Introduction:] \mbox{}\\ 
        In this part, a robot speaks ice break talks like self-introduction and explanations about sightseeing spots chosen by a participant.
        
    \item [\bfseries 2. a set of questions for collecting information:] \mbox{}\\ 
        After an ice-break talk in introduction part, a robot begins collecting the required information for the recommendation task.
        Hence, this part consists of some questions for a participant and the robot's responses against its answer.
        
    \item [\bfseries 3. Conclusion:] \mbox{}\\ 
        A robot, in this part, concludes a conversation with a participant, then a robot performs recommendation of sightseeing spots according to information about participants collected in part {\bfseries 2.}
\end{description}
Specifically, a robot that employed CF communicates with a participant aligned with a flow shown in Fig.~\ref{fig:cf-flow}.
Part {\bfseries 2.} has a self-loop-shaped arrow; this arrow means to select questions repeatedly as long as the left time for speaking conclusion part.
By modeling repeated selection at random, a robot can speak different contents of conversation every time a robot start-up.
In another case, CF enables a robot to select questions preferentially depending on a given recommendation task.

\begin{figure}[t]
    \centering
    \includegraphics[width=0.4\hsize]{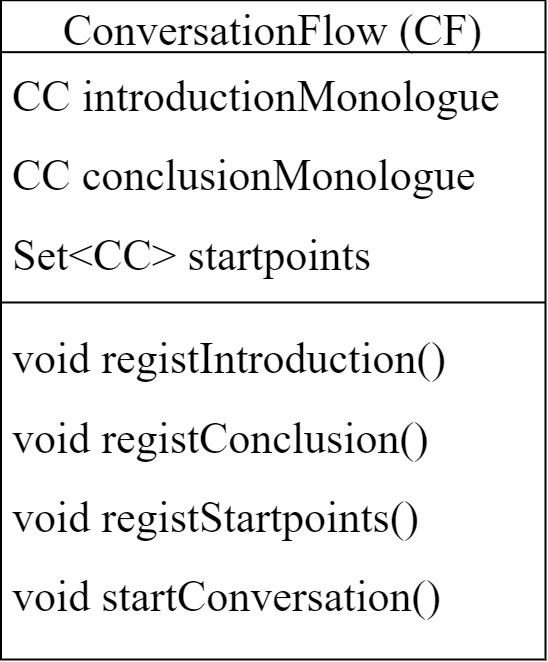}
    \caption{This class diagram partially displays ConversationFlow class's member variables and methods.}
    \label{fig:CF}
\end{figure}

Fig.~\ref{fig:CF} partially shows a design of CF class, where {\tt registIntrodcution()}, {\tt registStartpoints()}, and {\tt registStartpoints()} are member methods for registering conversations for each part in CF, and {\tt startConversation()} is for beginning conversations according with preregistered each part.

\begin{figure}[t]
    \centering
    \includegraphics[width=0.78\hsize]{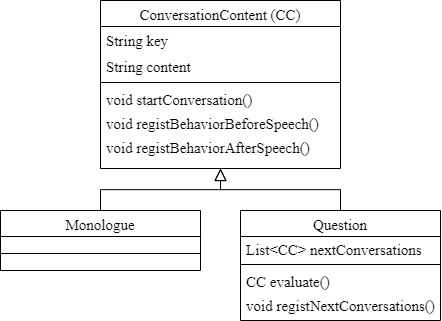}
    \caption{This class diagram partially displays ConversationContent class enabling us to define a content of a conversation. Monologue and Question class are sub-classes of ConversationContent class.}
    \label{fig:CC}
\end{figure}

CF's member variables {\tt introductionMonologue}, {\tt conclusionMonologue}, and {\tt startpoints} typed ``ConversationContent (CC)'' class as shown in Fig.~\ref{fig:CC}.
CC enables us to predefine specifically a content of conversations what a robot speaks or questions is.
Moreover, CC can register robot's gestures by invocating {\tt registBehaviorBeforseSpeech()} or {\tt registerBehaviorAfterSpeech()} so that a robot performs behaviors while speaking.
Furthermore, CC has two sub-classes as follows:
\begin{description}
    \item [\bfseries Monologue:] \mbox{}\\ 
        This sub-class enables us to define a content of what a robot unilaterally speaks to a person is.
        Hence, while a robot is speaking, microphones  off.
        
    \item [\bfseries Question:] \mbox{}\\ 
        This sub-class lets a robot question something to a person, so it requires a sentence of a question.
        At the same time, it needs to register two or more CCs which are a robot's line responding to a person's answer.
        When an instance of CC innovates {\tt evaluate()}, a CC selects a CC from registered next conversations ({\tt nextConversations}) by evaluating whether there is a value of CC's {\tt key} in an answer of a person.
        As we described above, this implementation is rule-base.
\end{description}

CC allows us to let a robot converse with a participant interactively by assembling predefined instances of {\tt Monologues} or {\tt Questions} in a tree structure.
A tree of CCs is a unit of a series of conversation, and we can register it into CF's {\tt startpoints} by invocating {\tt registStartpoints()}. 

\begin{figure}[t]
    \begin{tabular}{c}
      \begin{minipage}{1.00\hsize}
          \centering
          \includegraphics[width=0.4\hsize]{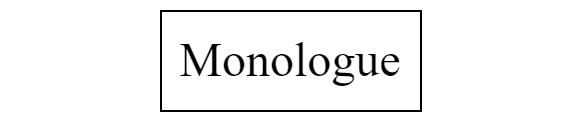}
          \subcaption{a unilateral speech}\label{sfig:unilateral-speech}
          \vspace{1ex}
      \end{minipage}
      \\
      \begin{minipage}{1.00\hsize}
          \centering
          \includegraphics[width=0.50\hsize]{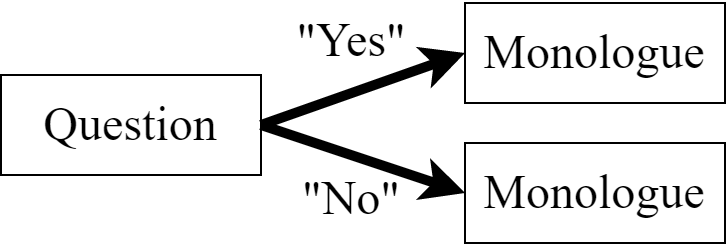}
          \subcaption{a yes-no (closed-ended) question}\label{sfig:yes-no}
          \vspace{1ex}
      \end{minipage}
      \\
      \begin{minipage}{1.00\hsize}
          \centering
          \includegraphics[width=0.50\hsize]{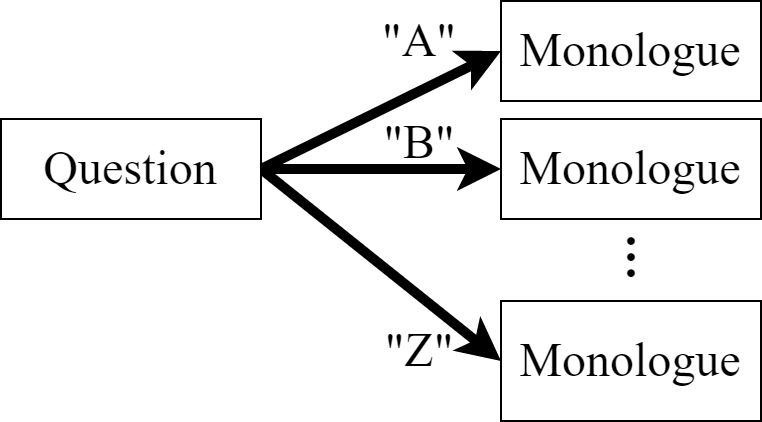}
          \subcaption{an open-ended question}\label{sfig:open-ended}
          \vspace{1ex}
      \end{minipage}
      \\
      \begin{minipage}{1.00\hsize}
          \centering
          \includegraphics[width=0.65\hsize]{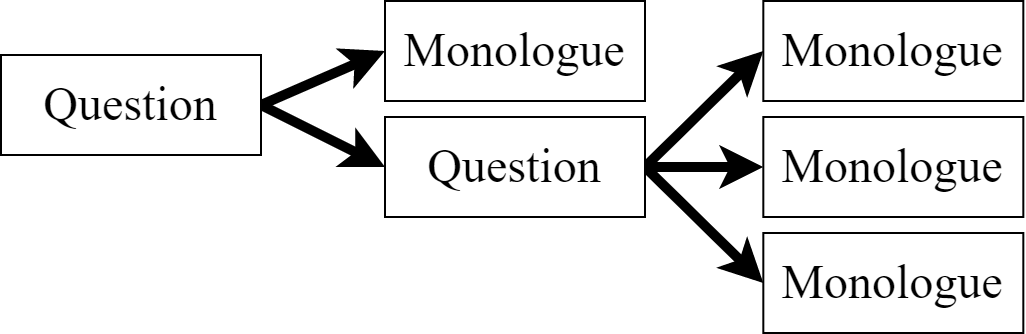}
          \subcaption{a multiple-step question}\label{sfig:multiple-step}
          \vspace{1ex}
      \end{minipage}
    \end{tabular}
    \caption{Four types of conversation constructed with CC}\label{fig:conv-group}
\end{figure}

Predefined conversations that can be constructed with CC are grouped into four types as shown in Fig.~\ref{fig:conv-group}.
A type of a conversation shown in Fig.~\ref{sfig:unilateral-speech} is a robot unilaterally speak to a person like an introduction or a conclusion-part in Fig.~\ref{fig:cf-flow}.
Three types of a conversation in Fig.~\ref{sfig:yes-no}, \ref{sfig:open-ended}, and \ref{sfig:multiple-step} are variety of a question for collecting person's information.
A yes-no question shown in Fig.~\ref{sfig:yes-no}, whose other name is an open-ended question, lets a person answer from limited options such as yes or no.
On the contrary, an open-ended question in Fig.~\ref{sfig:open-ended} enables a person to answer freely against the robot's question.
Finally, a multiple-step question in Fig.~\ref{sfig:multiple-step} realizes a series of questions that can change the content of the continuing question depending on a person's answer to a robot's question.

%% file: tex/030-cisrobot-definition.tex
\begin{figure*}
    \centering
    \includegraphics[width=\linewidth]{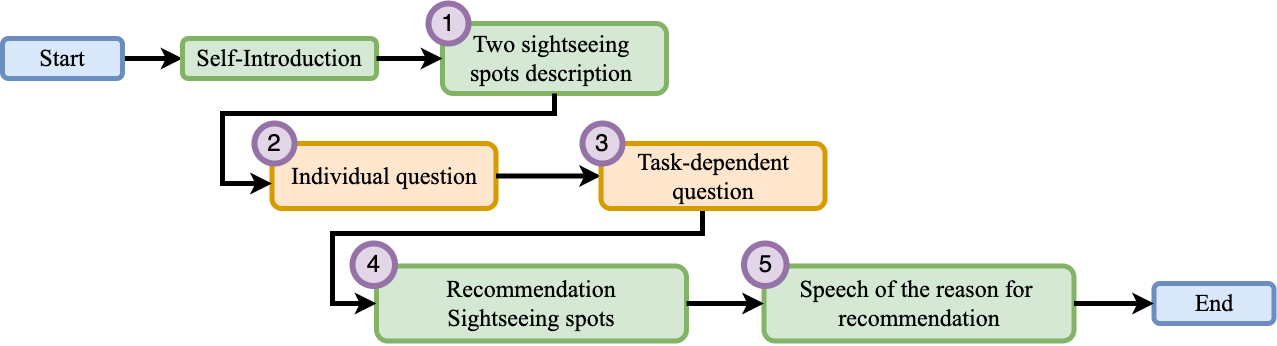}
    \caption{{\bfseries Overall dialogue flow: } The figure shows the flow of the entire conversation.The conversational flow consists of five elements (except for the self-introduction), and the purple numbers in the figure correspond to the five elements in section \ref{s:definition}.The colors of the boxes indicate the start and end of the dialog in blue, the robot's explanation in green, and the questions to the participants in yellow.
    }
    \label{f:conversation_flow}
\end{figure*}

In this Section, we describe the functional aspects of our dialog robot system. Specifically, Section \ref{ss:conv_flow} describes how a conversation with the robot takes place, and Section \ref{ss:non_verbal} describes the functionality of the non-verbal parts of the conversation.

\subsection{Conversation flow}\label{ss:conv_flow}
This section describes the flow of conversation in this dialog system.
The conversation flow can be roughly divided into the following five phases in Fig. \ref{f:conversation_flow}.

\begin{enumerate}
    \item Description of sightseeing spots
    \item Individual question
    \item Task-dependent question
    \item Sightseeing spot recommendations
    \item Speech of the reason for the  recommendation
\end{enumerate}

The dialogue is built around questions with scenarios and includes two types of questions, individual and task-dependent ones. 

\subsubsection{Description of sightseeing spot}
First, a description of the two recommended destinations is provided. The competition organizer has prepared a description of the tourist attractions, briefly explaining the facilities and location conditions. 

\subsubsection{Individual question}
Individual questions are task-independent inquiries related to personal attributes such as preferences, hobbies, and personality. 
Table \ref{t:q1toq3} shows the actual questions asked.

\subsubsection{Task-dependent question}
Task-dependent questions are used for sightseeing spot recommendations. 
The task-dependent questions are divided into two parts: the first half are questions that apply to sightseeing spots in general, and the second half are questions that are limited to sightseeing spots to be recommended, which are manually created based on API tags.
In this case, CISRobot has to recommend travel destinations so that we gather the user's opinions related to a short trip. Table \ref{t:q4toq6} shows the actual task-dependent questions asked. 

In the latter part, we ask questions corresponding to the tags we have created from tags obtained from the Google Places API \footnote{``Google Places API", \url{https://developers.google.com/maps/documentation/\\places/web-service},(accessed October 19, 2022)}. This tag is called ``place type” and contains a rough description of the location, such as ``amusement park” or 
``establishment.” For example, if a recommended sightseeing spot’s location type is ``amusement park,” the question is,`` Did you visit an amusement park recently?”. When making a recommendation, the user can provide reasons for the recommendation in this manner. Furthermore, the use of place types associated with the tourist attractions to be recommended allows for the asking of specific questions that task-dependent and individual questions do not allow for, as well as the preparation of questions for each place type, even if the number of tourist attractions is limited. 

\subsubsection{Sightseeing spot recommendations}
After the above questions, a sightseeing spot will be recommended.
The recommended sights are selected by a server set up by the competition.

\subsubsection{Speech of the reason for recommendation}
Following the above questions, a tourist attraction is suggested.
We asked the last question for each place type and one random question from among the other questions that had favorable answers for the recommendation as the decisive question for the recommendation after describing the recommended sightseeing spot.
By stating the basis for the recommendation in this manner, it is possible to give the user the impression that the recommended sightseeing spot was chosen based on the user's preferences.

Questions are prepared ahead of time, and they should be simple to answer and concise.
This is done to keep conversation breakdowns to a minimum.
Furthermore, responses to the questions were prepared ahead of time so that if an answer was not heard or was not prepared, a response of ``I see" was prepared so that the conversation was not interrupted.

\begin{table}[t]
    \centering
    \caption{Individual Question}
    \begin{tabular}{c|c}\hline
        Individual Questions & Description \\\hline
        Q1                 & Are you indoor or outdoor      \\
        Q2a                & What do you usually do at home \\
        Q2b                & Where do you usually go        \\
        Q3                 & Do you often take pictures    \\\hline
    \end{tabular}
    \label{t:q1toq3}
\end{table}

\begin{table}[t]
    \centering
    \caption{Task-dependent Question}
    \begin{tabular}{c|c}\hline
        Task-dependent Question & Description\\\hline
        Q4 & What transportation do you use                \\
        Q5 & Who will attend with you                      \\
        \multirow{2}{*}{Q6} & Which do you prefer  \\
        & experiencing or observing \\\hline
    \end{tabular}
    \label{t:q4toq6}
\end{table}

\subsection{Non-verbal behavior}\label{ss:non_verbal}
In this system, the robot's facial expressions and body movements were controlled as the robot's non-verbal behaviors during the dialog.

\subsubsection{Robot's facial expressions}
In this system, designed to serve customers at a travel agency, we used multiple friendly facial expressions because we thought that smiling facial expressions are essential to provide hospitality services. The ``Mood Base" and ``Full Smile" expressions described below are among the preset.
The ``Mood Base" was used as the default face for this dialog robot. This facial expression gives a soft impression to the user and does not give the robot characteristic stiffness.
The ``Full Smile" was used as the dialog robot's smile. This facial expression was simultaneously used when the robot introduced itself by saying its name and asking questions to the participants.
The parameters of this facial expression were set so that the degree of smiling would be more potent than that of ``Keep Smile," described below.
``Keep Smile" was created as a light smile for this dialog robot. This facial expression was created based on ``Mood Base" by changing the shape of the eyes and the corners of the mouth by adjusting four parameters: ``Valence," ``Arousal," ``Dominance," and ``Real intention."
By narrowing the shape of the eyes vertically and slightly raising the corners of the mouth, we tried to ease the participants' tension toward the robot and create a friendly impression. This facial expression was simultaneously executed in situations where the robot responded to the participant's answer with a sympathetic reply.
In addition, a system was also implemented in which the degree of the smile expression increased as the conversation with the participant progressed.
There were four stages, with the eye shape gradually becoming friendlier and the corners of the mouth rising. We believe that by implementing this system, it is possible to realize an exchange similar to a human conversation in which the tension is gradually dissolved.
\begin{table*}[ht]
    \centering
    \caption{Average score for each questionnaire item}
    \begin{tabular}{c|ccccccccc|c}\hline
        Questionnaire Number & i &ii&iii&iv&v&vi&vii&viii&ix& Total \\\hline
        CIS (our team) & \textbf{4.65} & \textbf{5.15} & 3.54 & \textbf{4.35} & \textbf{4.62} & \textbf{4.50} & 4.04 & \textbf{4.85} & \textbf{4.31} & \textbf{40.00}\\
        Team average & 4.48 & 4.46 & \textbf{3.74} & 4.32 & 4.60 & 4.45 & \textbf{4.35} & 4.83 & 4.23 & 39.47\\\hline
        Difference in points & +0.17 & \textbf{+0.69} & -0.20 & +0.02 & +0.02 & +0.05 & -0.31 & +0.01 & +0.08 & +0.53 \\
        \hline
    \end{tabular}
    \label{t:compe_res}
\end{table*}
\subsubsection{Robot's physical action}
In addition to facial expressions, the dialog robot used in this system can operate its upper body, allowing it to create postures such as head-swinging and forward leaning. In order to make human-like movements within the movable domain, several body movements were used in this system.
In a dialog robot's conversation with a human, body movements such as nodding and aids are essential to convey that the robot is listening to what the user is saying. Therefore, the system prepared several body movements for different nodding and aids and executed them as appropriate when the movements were required.

%% file: tex/040-experiment.tex
In this section, we describe the results of the actual interaction with the developed program. 
Section \ref{ss:results} describes the results of the post-conversation questionnaire and the team's results in DRC2022\cite{DRC2022}. 
Section \ref{ss:analysis} describes the results of a detailed analysis and further discussion using video recordings of the interaction.
Finally, a discussion of the competition results is provided in Section \ref{ss:discussion}.

%% file: tex/041-competition-result.tex
In this section, we describe the competition regulations and our team's results.
The competition was held over one day in August 2022 at the National Museum of Emerging Science and Innovation (Miraikan) in Tokyo, where many people interested in advanced technology gather.
The experience procedure is as follows.

\begin{table}[ht]
    \centering
    \caption{Description of questionnaire items}
    \begin{tabular}{c|c}\hline
        Questionnaire Number & Description  \\\hline
        i & Satisfaction with choice \\
        ii & Sufficiency of information \\
        iii & The nature of the dialogue \\
        iv & Adequacy of response \\
        v & Favorable response\\
        vi & Dialogue satisfaction\\
        vii & Robot reliability\\
        viii & Degree of information reference\\
        ix & Degree of desire to return \\\hline
    \end{tabular}
    \label{t:question_description}
\end{table}

\begin{enumerate}
    \item Participants answer how much they want to visit the two sightseeing spots selected by the competition organizer before the experience.
    \item Participants experience a dialogue with the robot.
    \item Participants answer a questionnaire after the dialogue.
\end{enumerate}

The dialogue system is executed as soon as the participants are seated, and the maximum duration of the dialogue is 5 minutes and 30 seconds.
The questionnaire that follows the dialogue is designed from two main perspectives.
The first is a questionnaire on the satisfaction of the interaction, and the second is a questionnaire on the expectations of the sightseeing spot.
The questionnaire has nine items to measure satisfaction with the interactive activities, and their descriptions are shown in Table \ref{t:question_description}.
The participants rated these items on a scale of 1 to 7, with one being the most satisfactory and seven being the least satisfactory.
Twenty-seven men and women in their teens to seventies experienced the dialogue, and 26 gave valid responses.

The results of these questionnaires are presented in Table \ref{t:compe_res}.
As can be seen from the bottom row that shows the difference in scores, many of the items were close to the average, with the highest score obtained for item ii.
Item ii is a questionnaire item that indicates whether the participants could hear enough information about sightseeing spots, and we assume that this is a result of our system's mechanism to answer many questions. As a result, the participants were able to ask many questions that they wanted to ask.
Items iii and vii, which were lower than the average, are items that evaluate the naturalness and reliability of the dialogue. It is possible that the feature of this system of asking many questions felt like an interrogation and affected these evaluations.

In addition, a questionnaire was administered to the participants to determine how much they wanted to visit each sightseeing spot after the dialogues.
This questionnaire was designed to measure changes in impressions of sightseeing spots before and after the dialogue and used a visual analog scale as the evaluation method.
The visual analog scale is a method using a seek bar with sightseeing spots on the left and right sides and can be evaluated in acceptable increments.
In this case, there were 100 steps between the two sightseeing spots, and all participants averaged the changes in these values before and after the dialogue.

The results are shown in Table \ref{t:compe_res_rec} and were slightly higher than the average of all teams.
Therefore, it is considered that the presentation of the rationale for the recommendation in this system has a positive effect on the effectiveness of the recommendation.
\begin{table}[ht]
    \centering
    \caption{Recommended effects of dialogue}
    \begin{tabular}{c|r}\hline
                       & score         \\\hline
        CIS (our team) & \textbf{13.44}         \\
        Team average & 13.09         \\
        \hline
    \end{tabular}
    \label{t:compe_res_rec}
\end{table}

%% file: tex/042-analysis.tex
In this Section, we describe a more detailed discussion using videos of the dialogues. Specifically, we will analyze the causes of this system's low evaluation of the naturalness of dialogue, as revealed by the results of the competition in Section \ref{ss:results}, regarding the presence or absence of dialogue breakdowns. 

First, annotations were made by watching the video to clarify whether there is a relationship between the breakdown of dialogue and the recommended effect. The case in which the respondent could not respond appropriately to a question and moved on to the next question was defined as a dialogue breakdown, and whether or not this occurred was counted as 0 or 1. The percentage of dialogue breakdowns among all questions was defined as the dialogue breakdown rate and was calculated for each participant. For example, if three out of the six questions resulted in dialogue failure, the dialogue failure rate would be 50 \%.

\begin{figure}[ht]
    \centering
    \includegraphics[width=\linewidth]{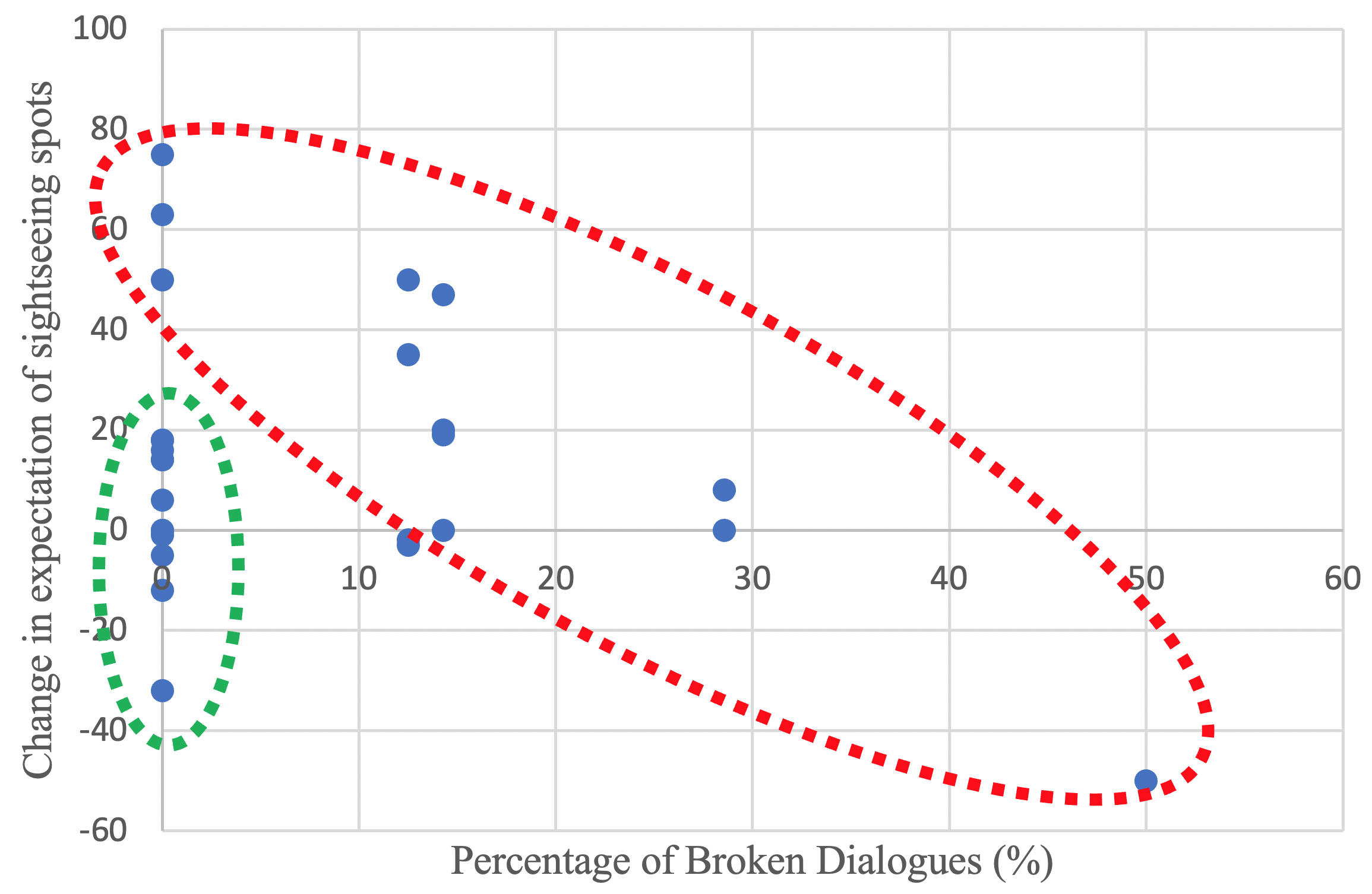}
    \caption{
    {\bfseries Scatter plots of each user's evaluation using the interactive video: }
    The vertical axis is the percentage of change in the impression of the sightseeing spot.
    The horizontal axis is the percentage of broken dialogues.
    }
    \label{fig:movie_analisys_result}
\end{figure}

The results are shown in Fig. \ref{fig:movie_analisys_result}. This figure plots the participants on the horizontal axis of the dialogue failure rate and the vertical axis of the degree they wanted to increase. Dialogue breakdowns occurred in 11 of the 23 participants, with an average breakdown rate of 8.24\% for each. The dotted oval in Fig. \ref{fig:movie_analisys_result} shows that the participants were divided into two groups. The area circled in red is the group for which the recommendation effect has decreased due to the breakdown of the dialogue. The area circled in red is the group in which it is evident that the breakdown of the conversation was the cause of the reduced recommendation effectiveness. On the other hand, the area circled in green is a group in which factors other than the breakdown are relevant since the effectiveness decreased even though the dialogue was established.

%% file: tex/043-discussion.tex
We also evaluated the dialogue videos qualitatively, looking at the negative points that stood out. 
As a result, it was found that the barge-in interrupted the other party's speech and that some people's facial expressions were ``scary" and did not give a good impression to some people. 

In summary, the following points can be made. First, the strong negative correlation between the dialogue breakdown rate and the recommendation effect reaffirms the need to reduce the dialogue breakdown rate. In addition, even among participants who did not experience dialogue breakdowns, the results were divided into two groups in recommendation effectiveness, suggesting that there are factors other than breakdowns related to recommendation effectiveness. One of these factors may be that the participants' facial expressions accompanying their utterances were inappropriate or did not give a natural impression due to barge-in.

%% file: tex/050-conclusion.tex
This paper describes developing a dialog robot system that recommends sightseeing spots based on dialogue. There are two significant issues to be addressed in the future. The first is to improve the accuracy of the recommendation basis. This is because the system randomly selects questions as the basis for its recommendations. However, it is necessary to construct a more accurate supporting system in the future. The second is to incorporate a generation-based system. In this study, the central systems, such as questions and responses, were created based on rules. Therefore, the cost of creating such a system is high when considering increasing the number of conversational variations. As mentioned in Section \ref{s:intro}, creating a consistent conversation using a generative model is challenging. However, there is a possibility that it can be used to generate a conversation in a part of the system. Therefore, it is necessary to consider whether a system that generates each element sequentially can be introduced into the conversation flow in the future.